\documentclass[conference]{IEEEtran}
\IEEEoverridecommandlockouts
\usepackage{cite}
\usepackage{amsmath,amssymb,amsfonts}
\usepackage{graphicx}
\usepackage[font=footnotesize]{subfig}
\usepackage{textcomp}
\usepackage{xcolor}
\usepackage{algorithm}
\usepackage[noend]{algpseudocode}
\usepackage{subfloat}
\usepackage[colorinlistoftodos]{todonotes}
\def\BibTeX{{\rm B\kern-.05em{\sc i\kern-.025em b}\kern-.08em
    T\kern-.1667em\lower.7ex\hbox{E}\kern-.125emX}}

\allowdisplaybreaks   

\newcommand{\ignore}[1]{}

\begin{document}

\title{Joint AP Probing and Scheduling: A Contextual Bandit Approach\\
}

\author{
    \IEEEauthorblockN{Tianyi Xu$^{a}$, Ding Zhang$^{b}$, Parth H. Pathak$^{b}$, Zizhan Zheng$^{a}$}
    \IEEEauthorblockA{$^a$ Department of Computer Science, Tulane University, New Orleans, LA, USA}
    \IEEEauthorblockA{$^b$ Department of Computer Science, George Mason University, Fairfax, VA, USA \vspace{-15pt}}
}

\ignore{
\author{\IEEEauthorblockN{1\textsuperscript{st} Tianyi Xu}
\IEEEauthorblockA{\textit{Computer Science Department} \\
\textit{Tulane University}}
\and
\IEEEauthorblockN{2\textsuperscript{nd} Ding Zhang}
\IEEEauthorblockA{\textit{Computer Science Department} \\
\textit{George Mason University}}
\and
\IEEEauthorblockN{3\textsuperscript{rd} Given Name Surname}
\IEEEauthorblockA{\textit{dept. name of organization (of Aff.)} \\
\textit{name of organization (of Aff.)}}
\and
\IEEEauthorblockN{3\textsuperscript{rd} Given Name Surname}
\IEEEauthorblockA{\textit{dept. name of organization (of Aff.)} \\
\textit{name of organization (of Aff.)}}
}
}
\newtheorem{theorem}{Theorem}
\newtheorem{claim}{Claim}
\newtheorem{remark}{Remark}
\newtheorem{corollary}{Corollary}
\newtheorem{lemma}{Lemma}
\newcommand{\red}[1]{{\color{red} #1}}
\newcommand{\blue}[1]{{\color{blue} #1}}
\newcommand\numberthis{\addtocounter{equation}{1}\tag{\theequation}}

\renewcommand{\baselinestretch}{0.93}


\maketitle

\begin{abstract}
 We consider a set of APs with unknown data rates that cooperatively serve a mobile client. The data rate of each link is i.i.d. sampled from a distribution that is unknown a priori. In contrast to traditional link scheduling problems under uncertainty, we assume that in each time step, the device can probe a subset of links before deciding which one to use. We model this problem as a contextual bandit problem with probing (CBwP) and present an efficient algorithm. We further establish the regret of our algorithm for links with Bernoulli data rates. Our CBwP model is a novel extension of the classic contextual bandit model and can potentially be applied to a large class of sequential decision-making problems that involve joint probing and play under uncertainty.
\end{abstract}


\section{Introduction}

Developing efficient resource allocation algorithms plays a central role in wireless networks research. Over the years, elegant solutions with performance guarantees have been designed for various tasks, including link scheduling, rate adaptation, power allocation, routing, and network utility optimization. A key assumption in many of these approaches is that the decision maker has perfect prior knowledge about the channel conditions. However, obtaining accurate channel conditions often requires substantial measurements, which can be very time consuming for both outdoor environments and dense indoor deployments. This is especially the case for mobile networks where the capacity of a link varies significantly over the locations of devices and environmental factors such as interference from other links and blockages. 
Thus, traditional approaches based on fixed channel conditions cannot obtain expected performance in unknown/uncertain environments or quickly adapt to changing environments. 

To cope with the various uncertainty in network resource allocation and obtain adaptive scheduling policies, learning based approaches have been intensively studied recently. In particular, various online learning based algorithms have been developed for link scheduling~\cite{modiano2019UCBgreedy,bo2019sleeping}, rate adaptation~\cite{atilla2019bandit} and beam selection~\cite{slivkins2011contextual}, just to name a few. These works consider the challenging setting where the capacity of a link follows an unknown distribution that can only be sampled when the link is activated (i.e., the bandit feedback). Instead of using an offline learning approach with separated data collection and decision making stages, they adopt a multi-armed bandit based online learning framework that integrates exploration and exploitation. By carefully balancing the two aspects, they obtain no-regret adaptive policies with long-run performance approaching what can be achieved by the best offline policies that have prior knowledge on channel conditions.

In many real settings, a decision maker may obtain additional observations beyond the pure bandit feedback. For example, in the next generation millimeter-wave 802.11ad/ay WLANs, beamforming can be used to infer the real-time link quality before a scheduling decision is made. However, beamforming between all APs and clients can be time consuming for a densely deployed WLAN. Thus it is more realistic to assume that only a subset of APs can be selected for beamforming in each round, which reduces channel uncertainty but does not completely mitigate it. 
In this case, it is crucial to jointly optimize AP selection for beamforming (probing) and link scheduling for serving clients (play). 

In this paper, we present a novel extension of the bandit learning framework to incorporate joint probing and play. We assume that before the decision maker chooses an arm to play in each round, it can probe a subset of arms and observe their rewards (in that round). The decision maker then picks an arm to play according to the observations obtained in the probing stage and historical data. Our framework can be directly applied to the joint beamforming and scheduling problem when multiple APs collaboratively serve a single client (detailed system model in Sec.~\ref{sec:model}). Given that the data rate a client can obtain from an AP is highly correlated with the client location, we consider a contextual bandit model and treat the client location (or an approximation of it) as the context and learn a context-dependent joint probing and play policy. 

To solve the problem, we first derive useful structural properties of the offline optimal solution and then develop an online learning algorithm by extending the contextual zooming algorithm in~\cite{slivkins2011contextual}. We further establish the regret bound of our algorithm in the special case when the reward distributions are Bernoulli. We apply our framework to the joint beamforming and scheduling problem in 802.11ad WLANs where a set of APs collaboratively serve a single mobile client. Simulations using real data traces demonstrate the efficiency of our solution.

Our bandit learning model and its extensions can potentially be applied to a large body of sequential decision making problems that involve joint probing and play under uncertainty. For example, by integrating probing with combinatorial multi-armed bandits where the decision maker can pick multiple arms to play, we can model the joint beamforming and scheduling problem in the more general multi-AP multi-client setting. 
As another example, consider the problem of finding the shortest path between a source and a destination in a road network with unknown traffic, where a path searcher can query a travel server to obtain hints of real time travel latency~\cite{bhaskara2020adaptive}. 
Since each query consumes server resources and incurs delay, the path searcher can only make a limited number of queries before picking a path. Further, the path searcher may utilize contextual information such as the current time to assist decision making. This problem can again be modeled as a contextual combinatorial bandit problem with probing. 

\ignore{
Our main contributions can be summarized as follows. 
\begin{itemize}
    \item We propose contextual bandit with probing (CBwP) as a novel bandit learning framework that incorporates joint probing and play into sequential decision making under uncertainty.
    \item We derive structural properties of the optimal offline solution and develop an efficient online learning algorithm to CBwP. We further derive the regret bound of the online algorithm when the rewards are Bernoulli.  
    \item We apply our framework to the joint beamforming and scheduling problem in 802.11ad WLANs where a set of APs collaboratively serve a single mobile client. Simulations using real data traces demonstrate the efficiency of our solution.
\end{itemize}
}

\section{Related work}
The classic multi-armed bandit (MAB) model provides a clean framework for studying the exploration vs. exploitation tradeoff in sequential decision making under uncertainty. Since the seminal work of Lai and Robbins~\cite{lai1985asymptotically}, MAB and its variants have been intensively studied~\cite{auer2002finite,gittins2011multi,bubeck2012regret} and applied to various domains including wireless resource allocation. 
In particular, a combinatorial sleeping multi-armed bandit model with fairness constraints is considered in~\cite{bo2019sleeping}, which has been used to model single AP scheduling where multiple clients compete for sending packets to the common AP. 
In~\cite{atilla2019bandit}, the problem of link rate selection for a single wireless link is considered and a constrained Thompson sampling algorithm is developed to exploit the structural property that a higher data rate is associated with a lower transmission success probability. 
In~\cite{modiano2019UCBgreedy}, online learning based scheduling for general ad hoc wireless networks with unknown channel statistics is considered. The classic greedy maximal matching based algorithm is extended by using UCB-based link weights. 
The work that is closest to ours is \cite{fml2018}, where a contextual multi-armed bandit algorithm is applied to the beam selection problem in mmWave vehicular systems. 
However, none of the above work considers the joint probing and scheduling problem as we consider in this paper.  

Probing strategies for independent distributions have been studied in various domains including database query optimization~\cite{munagala2005pipelined,liu2008near,deshpande2016approximation} and wireless communication~\cite{guha2006optimizing}. 
A common setting is that given a set of random variables with {\it known} distributions, a limited number of probes (observations) can be made about these distributions. A selection decision is then made according to the observations. This corresponds to the offline problem in our setting. Various objective functions have been considered including maximizing the largest value found minus the total probing cost spent. Since the general problem is NP-hard, various approximation algorithms have been developed~\cite{goel2006asking}. More recently, adaptive probing strategies have been studied for shortest path routing~\cite{bhaskara2020adaptive} and when the probing cost varies across different alternatives (i.e., arms)~\cite{attias2017stochastic}. 
However, these results do not apply to the online settings with {\it unknown} distributions considered in this paper.

\section{System Model and Problem Formulation}\label{sec:model}

In this section, we define contextual bandits with  probing (CBwP) as a novel extension of the classic contextual bandits model \cite{slivkins2011contextual}. To make it concrete, we use the joint AP probing and selection problem as an example when presenting the model. Our formulation applies to a large class of sequential decision problems that involve joint probing and selection under uncertainty. We further derive some important properties of CBwP. 

\subsection{Contextual Bandits with Probing (CBwP)}

We consider a set of APs connected to a high-speed backhaul that collaboratively serve a set of mobile clients. AP collaboration helps boost wireless performance in both indoor and outdoor environments and is especially useful for directional mmWave communications that are susceptible to blockage~\cite{fml2018}. For simplicity, we assume that the beamforming process determines the best beam (i.e., highest SNR) from AP to the client. Hence, we do not distinguish AP selection from beam selection. Our framework readily applies to the more general setting of joint AP and beam selection. 

To simplify the problem, we focus on the single client setting in this work. 
Let $X$ be a set of contexts that correspond to the location (or a rough estimate of it) of a moving client. In general, $X$ can be either discrete or continuous. Let $A$ be a discrete set of arms that correspond to the set of APs, and $N \triangleq |A|$. We consider a fixed time horizon $T$ that is known to the decision maker. In each time step $t$, the decision maker first receives a context $x_t\in{X}$ and then plays an arm $a_t\in{A}$ and receives a reward $\phi(a_t|x_t) \in [0,1]$, which is $i.i.d.$ sampled from an {\it unknown} distribution, $\Phi(a_t|x_t)$, that depends on both the context $x_t$ and the arm $a_t$. 
We assume that the expected value of $\Phi(a|x)$ exists for any context-arm pair $(x,a)$ and denote it by $\mu(a|x)$. The sequence of context $(x_t)_{t \in \mathbb{N}}$ is assumed to be external to the decision making process. In the AP selection problem, the reward corresponds to the data rate that a client at a certain location can receive from an AP. 


In the classic contextual bandit problem, the instantaneous reward of an arm is revealed only when it is played, and the decision maker receives no side observations. In contrast, we consider a more general setting where after receiving the context, the decision maker can first probe a subset of $K < N$ arms and observe their rewards, and then pick an arm to play (which may be different from the set of probed arms). In general, probing an arm reduces the uncertainty about the arm. We assume that the probing period (for $K$ arms) is short enough so that if an arm $a$ is probed with $\phi(a|x_t)$ observed, then the same $\phi(a|x_t)$ is the reward obtained if arm $a$ is played in $t$. 
However, probing does reduce packet transmission time; hence, we require $K$ to be relatively small. The problem of choosing a proper $K$ either statically or dynamically is left to our future work.
We further assume that the probing results are independent across arms. That is, $\phi(a|x_t)$ is independent of other arms probed in $t$ or before $t$. Extension to correlated arms is left to future work. 





Let $G_{\pi_t}(x_t)$ denote the {\it expected} reward in time step $t$ under a (time-varying) joint probing and play policy $\pi_t$, where the expectation is over the randomness of observations in time step $t$. Similar to the classic contextual bandit model, our goal is to maximize the (expected) cumulative reward 
$\mathbb{E}[\sum_{t=1}^TG_{\pi_t}(x_t)]$.
As we discuss below, when the reward distribution $\Phi(a|x)$ is known a prior for each context-arm pair $(x,a)$, the single stage problem at each time step can be modeled as a Markov decision process with an optimal policy. Let $G^{\star}(x_t)$ denote the expected reward under $x_t$ when the optimal {\it offline} policy is adopted in each time step. Define the total regret as follows: 
\begin{align}
R(T)=\sum_{t=1}^T (G^{\star}(x_t) - G_{\pi_t}(x_t))
\end{align}
The goal of maximizing the expected cumulative reward then converts to minimizing $\mathbb{E}(R(T))$. 

Similar to \cite{slivkins2011contextual}, we assume that the context set $X$ is associated with a distance metric $\mathcal{D}$ such that $\mu(a|x)$ satisfies the following Lipschitz condition:
\begin{align}
|\mu(a|x)-\mu(a|x')| \leq \mathcal{D}(x,x'), \forall a \in A, x, x' \in X
\label{Lipschitz}
\end{align}
Without loss of generality, we assume that $\mathcal{D}(\cdot,\cdot) \leq 1$. This condition helps us capture the similarity between the context-arm pairs. In the joint AP probing and selection problem, $\mathcal{D}$ is defined as the Euclidean distance between locations. 

\ignore{
Similar to \, we assume that there is a distance metric $D$ over $X$ such that the metric space $(X,D)$ is a similarity space that satisfies the following Lipschitz condition:
\begin{align}
|\mu(a|x)-\mu(a|x')| \le D(x, x')=\rm{min}(1, \it{L|x-x'}|), \forall a \in A \end{align}
where $L$ is the normalization of the distance difference between $x$ and $x'$. Without loss of generality, we assume that $D(x,x') \le 1$ for any $x,x' \in X$.
}




\subsection{Offline Problem as an MDP}
We first consider the offline setting where the reward distributions are known to the decision maker a prior. We show that the joint probing and play problem in each time step can itself be modeled as a Markov decision process (MDP). We further derive important properties of the MDP. 

\ignore{Due to the space limitation, we omit all the proofs. The reader is referred to~\cite{tech} for the missing details.}

Consider any time step with a context $x$. To simplify the notation, we omit the time step subscript in this section. At each probing step $i \in \{0,1,...K\}$, the decision maker observes the current state $s_i \triangleq (a_1,\ldots,a_i,\phi(a_1|x),\ldots, \phi(a_i|x))$ and then chooses the next arm $a_{i+1}$ to probe, where $a_j$ is the arm probed in round $j$ and $\phi(a_j|x)$ is the observed reward of arm $a_j$ under context $x$. We define $s_0 \triangleq \emptyset$. Further, the decision maker can decide at any round $i \leq K$ to stop probing and pick an arm to play according to the probing result, and receives the reward of the played arm. We observe that more information always helps in our problem, thus it never hurts to wait until round $K$ to choose an arm to play. Let $S$ denote the set of all possible states and $\mathcal{P}(A)$ the set of distributions over $A$. The joint problem can be solved using a pair of policies: a probing policy $\pi_1: S \rightarrow \mathcal{P}(A)$ that maps an arbitrary probing history to the next arm to probe and a play policy $\pi_2: S \rightarrow \mathcal{P}(A)$ that chooses an arm to play according to the probing result. Let $\pi=(\pi_1,\pi_2)$ denote a joint probing and play policy. 

We first observe that in the offline setting, there is a simple deterministic play policy that is optimal. Let $g_{\pi_2}(s_i)$ denote the expected reward that can be obtained from playing an arm using policy $\pi_2$ given the probing result $s_i$ after $i$ rounds. We have
\begin{align*}
g_{\pi_2}(s_i)=\sum_{j=1}^{i}{\pi_2}(a_j|s_i)\phi(a_j|x)+\sum_{b_j \in {A \backslash \{a_1,...,a_i\}}}\hspace{-4ex}{\pi_2}(b_j|s_i)\mu(b_j|x)
\end{align*} 
where $\pi_2(a|s)$ denotes the probability of playing arm $a$ given the probing result $s$. For any arm $a$, let $v(a|x,s_i)=\phi(a|x)$ if $a \in \{a_1,...,a_i\}$ and $v(a|x,s_i)=\mu(a|x)$ otherwise. Then we observe that the deterministic policy that always plays an arm with maximum $v(a|x,s_i)$ is optimal and obtains the following optimal reward:
\begin{align}
g^\star(s_i)={\rm{max}}\Big\{\mathop{\rm{max}}\limits_{a\in\{a_1,\ldots,a_i\}}\phi(a|x_t),\mathop{\rm{max}}\limits_{b\in{A \backslash \{a_1\ldots,a_i\}}}\mu(b)\Big\}
\label{eq:optimal_play_reward}
\end{align}
We summarize this observation as a lemma:
\begin{lemma}
Given any context $x$ and state $s_i$, the deterministic policy that plays an arm with maximum $v(a|x,s_i)$ is optimal.
\label{lemma:optimal-play}
\end{lemma}

We then consider the problem of finding an optimal probing policy. For any given play policy $\pi_2$, the probing problem can be formulated as a finite-horizon MDP $M= (S, A, rwd, tr, K)$, where $S$ is a set of states defined above, $A$ is set of actions that correspond to the set of arms. 
The reward function $rwd: S \times A \times S \rightarrow \mathbb{R}$ is defined as 
$rwd(s_i,a_{i+1},s_{i+1})=g_{\pi_2}(s_{i+1})-g_{\pi_2}(s_i)$ for $i<K$ and $rwd(s_i,a_{i+1},s_{i+1})=0$ otherwise. The transition dynamics $tr(s_{i+1}|s_i,a_{i+1})$ gives the probability of reaching state $s_{i+1}$ given the current state $s_i$ and action $a_{i+1}$, which can be derived from ${\rm{Pr}}(s_{i+1}=(s_i,a_{i+1},\phi(a_{i+1}|x))|s_i,a_{i+1})={\rm{Pr}}(\Phi(a_{i+1}|x)=\phi(a_{i+1}|x))$. 

We consider the standard objective of maximizing the expected cumulative reward for the MDP. Given the way the reward function is defined, this can be represented as $G_{\pi} \triangleq \mathop{\mathbb{E}}_{\pi,\Phi}[\sum_{i=1}^{K-1} rwd(s_i,a_{i+1},s_{i+1})] = \mathop{\mathbb{E}}_{\pi_1,\Phi}[g_{\pi_2}(s_K)]$. 
Thus, to find the optimal $\pi$, it suffices to adopt an optimal play policy  $\pi^{\star}_2$ (such as the deterministic policy defined above) and solve the MDP to find the optimal probing policy $\pi^{\star}_1$. Let $\pi^{\star}=(\pi^{\star}_1,\pi^{\star}_2)$ denote the optimal joint (offline) policy.

 The MDP $M$ defined above uses the complete history of the probing results as the state. We then show that assuming an optimal play policy is adopted, it suffices to keep the set of arms probed and the maximum reward observed. This allows us to obtain a smaller MDP without loss of optimality. To show this, given any state $s_i = (a_1,...,a_i,\phi(a_1|x),...,\phi(a_i|x))$, we derive a new state $\overline{s}_i \triangleq (a_1,...,a_i,\max(\phi(a_1|x),...,\phi(a_i|x)))$. Let $\overline{S}$ denote the set of states $\overline{s}$. We further say that $s$ is similar to $\overline{s}$ (denoted by $s \sim \overline{s}$) if the latter can be derived from the former. We then define a new MDP $M'=(\overline{S}, A, rwd', tr',K)$, where $rwd'(\overline{s}_i,a_{i+1},\overline{s}_{i+1}) = g^{\star}(s_{i+1})-g^{\star}(s_i)$ for any $s_i$ and $s_{i+1}$ such that $s_i \sim \overline{s}_i$ and $s_{i+1} \sim \overline{s}_{i+1}$. Note that the reward function is well defined as $g^{\star}(s_i)$ only depends on the maximum probed value in $s_i$ (see Equation~\eqref{eq:optimal_play_reward}). Further, the new transition dynamics $tr'$ can be derived from the following observation:
 \begin{align}
{\rm{Pr}}(\overline{s}_{i+1}|\overline{s}_i,a_{i+1})=\mathop{\rm \sum}\limits_{s_{i+1} \sim \overline{s}_{i+1}}{\rm{Pr}}(s_{i+1}|s_i,a_{i+1}), \forall s_i \sim \overline{s}_{i}. 
\label{eq:transition}
\end{align}

We then show that $M$ and $M'$ have the same optimal value. Let $Q^{\star}_M(s_i,a_{i+1}) \triangleq \max_\pi \mathbb{E}_{\pi,\Phi}[\sum^{K-1}_{j=i}rwd(s_j,$ $a_{j+1},s_{j+1})|s_i,a_{i+1}]$ denote the optimal state-action value function of $M$ for any state $s_i$ and action $a_{i+1}$. Assuming $Q^{\star}_M(s_K,a_{K+1}) =0$, $Q^{\star}_M$ satisfies the Bellman optimality equation:  
\begin{align*}
Q_M^\star(s_i,a_{i+1})=&\mathop{\rm \sum}\limits_{s_{i+1} \in S}{\rm{Pr}}(s_{i+1}|s_i,a_{i+1})\big[rwd(s_i,a_{i+1},s_{i+1})\\
&\hspace{18ex}+\mathop{\rm max}\limits_{a' \in A}Q_M^\star(s_{i+1},a')\big]
\end{align*}
$Q_{M'}^\star(\overline{s}_i,a_{i+1})$ is defined analogously. We then have the following result, which can be derived using the Bellman optimality equation and mathematical induction:  

\begin{lemma}
$Q_M^\star(s_i,a_{i+1})=Q_{M'}^\star(\overline{s}_i,a_{i+1}), \ \ \forall s_i \sim \overline{s}_i$.
\label{lemma1}
\end{lemma}


\begin{IEEEproof}
We prove the result by mathematical induction. First, we have $Q_M^\star(s_{K-1},a_{K})=Q_{M'}^\star(\overline{s}_{K-1},a_K) = 0$. Thus, the base case holds. Assume the result holds for $i+1$. From the Bellman optimality equation of $Q_{M'}^\star$, we have
\begin{align*}
Q_{M'}^\star(\overline{s}_i,a_{i+1})=&\mathop{\rm \sum}\limits_{\overline{s}_{i+1}}{\rm{Pr}}(\overline{s}_{i+1}|\overline{s}_{i},a_{i+1})\big[rwd'(\overline{s}_{i},a,\overline{s}_{i+1})\\
&\hspace{18ex}+\mathop{\rm max}\limits_{a'}Q_{M'}^\star(\overline{s}_{i+1},a')\big] \\
\overset{\text{(a)}}=&\mathop{\rm \sum}\limits_{\overline{s}_{i+1}}\mathop{\rm \sum}\limits_{s_{i+1} \sim \overline{s}_{i+1}}\hspace{-2ex}{\rm{Pr}}(s_{i+1}|s_i,a_{i+1})\big[rwd'(\overline{s}_{i},a,\overline{s}_{i+1})\\
&\hspace{18ex}+\mathop{\rm max}\limits_{a'}Q_{M'}^\star(\overline{s}_{i+1},a')\big] \\
\overset{\text{(b)}}=&\mathop{\rm \sum}\limits_{\overline{s}_{i+1}}\mathop{\rm \sum}\limits_{s_{i+1} \sim \overline{s}_{i+1}}\hspace{-2ex}{\rm{Pr}}(s_{i+1}|s_i,a_{i+1})\big[rwd(s_{i},a,s_{i+1})\\
&\hspace{18ex}+\mathop{\rm max}\limits_{a'}Q_M^\star(s_{i+1},a')\big]\\
=&\mathop{\rm \sum}\limits_{s_{i+1}}{\rm{Pr}}(s_{i+1}|s_i,a_{i+1})\big[rwd(s_{i},a,s_{i+1})\\
&\hspace{18ex}+\mathop{\rm max}\limits_{a'}Q_M^\star(s_{i+1},a')\big] \\
=&Q_{M}^\star(s_i,a_{i+1}).
\end{align*}

\noindent where (a) follows from~\eqref{eq:transition} and (b) follows from the definition of the reward function $rwd'$ and the inductive hypothesis. 
\end{IEEEproof}

\subsection{Offline Problem with Bernoulli Rewards}
When $\Phi(a|x)$ follows a Bernoulli distribution (fully defined by $\mu(a|x)$) for any context-arm pair $(x,a)$. 
there is a simple non-adaptive probing policy that is optimal. 

\begin{lemma}
Consider the following non-adaptive probing policy for arms with Bernoulli rewards: given any context $x$, find the $K+1$ arms with the largest $\mu(a|x)$ among all the arms, and then probe any $K$ of them. This policy together with the deterministic play policy given in Lemma~\ref{lemma:optimal-play} gives an optimal joint policy to the offline problem.

\label{lemma:optimal-policy}
\end{lemma}

\begin{IEEEproof}
We first observe that for arms with Bernoulli rewards, there is an optimal joint probing and play policy that has the following form: At state $s_0$, probe an arm $a_1$. If $\phi(a_1|x)=1$, play $a_1$ and stop. Otherwise, probe another arm $a_2$. Repeat this process until a probed arm gives a reward of 1 or $K$ arms have been played. In the latter case, play an arm $a_{K+1}$ that has not been probed. 

Consider any policy $\pi$ of the above form. Let $\{a_1, ..., a_K\}$ be the set of arms involved in the policy. We construct a new joint policy $\pi'$ where we first probe all the $K$ arms $\{a_1, ..., a_K\}$ involved in $\pi$ and then play an arm according to the optimal deterministic policy. We observe that the expected reward of $\pi'$ is no less than that of $\pi$. Thus, there is a non-adaptive probing policy that is optimal. Further,  
\begin{align*}
G_{\pi'}=&\left(1-\prod\limits_{i=1}^{K}(1-\mu(a_i|x))\right)\\
&+\prod\limits_{i=1}^{K}(1-\mu(a_i|x))\max_{b \in A \backslash \{a_1,...,a_K\}}\mu(b|x)\\
=&1-\prod\limits_{i=1}^{K}(1-\mu(a_i|x))(1-\max_{b \in A \backslash \{a_1,...,a_K\}}\mu(b|x))
\end{align*}
Thus, $G_{\pi'}$ is maximized when the set of arms $a_1,...,a_K$ and $b \in A \backslash \{a_1,...,a_K\}$ have the maximum expected reward $\mu(\cdot|x)$ among all the arms, irrespective of the order that these arms are probed and played.
\end{IEEEproof}

\section{The CBwP Algorithm}\label{sec:algo}

Lemma~\ref{lemma:optimal-policy} indicates that in the offline setting, greedy probing plus greedy play is optimal for Bernoulli rewards. However, this approach cannot be directly applied to the online setting as $\Phi(a|x)$'s are unknown, which involves a fundamental exploration vs. exploitation tradeoff. In this section, we consider the online setting and design an algorithm for the CBwP problem. We further derive its regret for the special case with Bernoulli rewards. 

\subsection{Algorithm Description}
Our algorithm is based on the contextual zooming algorithm in \cite{slivkins2011contextual}, which adaptively partitions the similarity space to exploit the Lipschitz condition 
(Equation~\eqref{Lipschitz}). 
As we consider a finite set of arms in our problem, we apply adaptive partitioning to the context space only. The main contribution of our work is extending the contextual zooming technique to the joint probing and play setting, which brings new challenges in both algorithm design and analysis.  

The algorithm (see Algorithm~\ref{alg:cbwp}) maintains a finite set $\mathcal{A}_a$ of active balls for each arm $a$. We require that the balls in $\mathcal{A}_a$ collectively cover the similarity space $(X,\mathcal{D})$. Initially $\mathcal{A}_a$ contains a single ball of radius 1. A ball is activated once it is added to $\mathcal{A}_a$ and remains active. These balls correspond to a partition of the context space from the arm $a$'s perspective. 

In each time round $t$, a context $x_t$ is revealed, and the algorithm selects up to $K$ arms $\{a_1,a_2,\ldots,a_K\}$ to probe according to the ``probing rule''. In each probing step after observing $\phi(a_i|x_t)$, the algorithm may activate a new ball according to the ``activation rule''. The probing stage terminates if either $K$ arms have been probed or $\phi(a_i|x_t) = 1$ for some $i$. The algorithm then selects an arm $b$ to play according to the ``playing rule", and may activate a new ball according to the ``activation rule''. 



We then specify the three rules used in the algorithm. Both the probing and play rules are inspired by Lemma~\ref{lemma:optimal-policy}. Since the true distributions of rewards are unknown, the algorithm picks arms according to their estimated rewards together with a confidence term. Let $r(B)$ denote the radius of a ball $B$. The {\it confidence radius} of $B$ at time round $t$ is defined as:
\begin{align}
{\rm conf}_t(B) \triangleq 4\sqrt{\frac{{\rm log}T}{1+n_t(B)}} \label{eq:conf}  
\end{align}
where $n_t(B)$ is the number of times $B$ has been selected from $t=1$ to $t$. Let ${\rm re}_t(B)$ denote the total reward from all rounds up to $t-1$ in which $B$ has been selected, and $v_t(B) \triangleq \frac{{\rm re}_t(B)}{{\rm max}(1, n_t(B))}$ the average reward from $B$. 
The {\it pre-index} of $B$ is defined as
\begin{align}
I_t^{\rm pre}(B) = v_t(B) + r(B) + {\rm conf}_t(B) 
\label{preUCB}
\end{align}
The {\it index} of $B \in \mathcal{A}_a$ is obtained by taking a minimum over all active balls of AP $a$:
\begin{align}
I_t(B) \triangleq r(B)+\mathop{\rm min}\limits_{B'\in \mathcal{A}_a}(I_t^{\rm pre}(B')+\mathcal{D}(B, B')), \forall B \in \mathcal{A}_a 
\label{UCB}
\end{align}
where $\mathcal{D}(B_a, B_a')$ is the distance between the centers of the two balls.

In each round $t$ and for each AP $a$, let $\mathcal{B}_a \subseteq \mathcal{A}_a$ be the set of active balls that contains $x_t$ and has the minimum radius. Let $B^{\rm{sel}}_{a}$ be an arbitrary ball in $\mathcal{B}_a$ with maximal index. 
We then state the three rules as follows:

\begin{itemize}
\item{\textbf{probing rule}}: At each probing step $i$ in time round $t$, the algorithm probes an arm $a_i$ with the maximal index $I_t(B^{\rm{sel}}_{a})$  (break ties arbitrarily) among the set of unprobed arms 
and gets observation $\phi(a_i|x_t)$. The probing stage ends if $K$ arms have been probed or $\phi(a_i|x_t)=1$ for some $i$.
\item{\textbf{playing rule}}: In each time round $t$ and after the probing stage is done, let $v(a) = \phi(a|x_t)$ if $a$ has been probed and $v(a) = I_t(B^{\rm{sel}}_{a})$ otherwise. If $\phi(a_i|x_t)=1$ for some $i$ in the probing stage, play $a_i$. Otherwise, play an arm $b$ with maximal $v(b)$. 
\item{\textbf{activation rule}}: If arm $a$ is probed or played in time round $t$, the algorithm updates $n_t(B^{\rm{sel}}_{a})$ and $\text{conf}_t(B_a)$. 
Further, a new ball with center $x_t$ and radius $\frac{1}{2}r(B^{\rm{sel}}_{a})$ is activated if $\text{conf}_t(B^{\rm{sel}}_{a}) \le$ $r(B^{\rm{sel}}_{a})$. $B^{\rm{sel}}_{a}$ is called the parent ball of this newly activated ball. 
\end{itemize}
\renewcommand{\algorithmicrequire}{\textbf{Input:}}  
\renewcommand{\algorithmicensure}{\textbf{Output:}} 
\begin{algorithm}[t]
  \caption{
  Contextual zooming for joint probing and play} 
  \label{alg:cbwp}
  \begin{algorithmic}[1]
    \Require
      $A$: a set of $N$ arms; $(X, \mathcal{D})$: a similarity space of diameter $\le
      $ 1; $T$: time horizon
       \For {each AP $a$}
       \State $B$ $\leftarrow$ $B(x, 1)$     \hspace*{1.5em}//center $x \in X$ is arbitrary
       \State $\mathcal{A}_a \leftarrow \{B\}$, \ $n(B)$ $\leftarrow$ $0$, \ re$(B)$ $\leftarrow$ $0$
       \EndFor
      \For{each round $t=1,2,\ldots,T$ }
      \State Input context $x_t$
      \For {each AP $a$}
      \State $\mathcal{B}_{a}$  $\leftarrow$ $\{B \in \mathcal{A}_a: r(B) = \mathop{\rm min}\limits_{B' \in \mathcal{A}_a, x_t \in B'} r(B')$ $\rm {\}}$
      \State $B^{\rm{sel}}_{a} \leftarrow$  $\mathop{\rm argmax}\limits_{B \in{\mathcal{B}_{a}}} {\it I_t(B)}$
      \State $v(a) \leftarrow I_t(B^{\rm{sel}}_{a})$
      \EndFor
      \For {$i=1$ to $K$} 
      \State $a_i \leftarrow \mathop{\rm argmax}\limits_{a_i'\in{A \backslash \{a_1,a_2,\ldots,a_{i-1}\}}} {\it I_t(B^{\rm{sel}}_{a'_i})}$ \hspace*{0.1em}//{\rm Probing} 
      \State \rm Probe $a_i$, get the observation $\phi(a_i|x_t)$
      \State $v(a_i) \leftarrow \phi(a_i|x_t)$
      \State \Call {Activation}{$\phi(a_i|x_t), B^{\rm{sel}}_{a_i}, x_t$}
      \If {$\phi(a_i|x_t)=1$} 
      \State Break
      \EndIf
      \EndFor
      \If{$\phi(a_i|x_t) = 1$ for some $i$}  \hspace*{1.5em}//{\rm Playing}
      \State Play arm $a_i$, get the reward 1
      \Else 
      \State $b \leftarrow \mathop{\rm argmax}\limits_{b\in{A}} v(b)$
      \State \rm Play arm $b$, get the reward $\phi(b|x_t)$
      \State \Call 
      {Activation}{$\phi(b|x_t), B^{\rm{sel}}_{b}, x_t$}
      \EndIf
     \EndFor
     \Function {Activation}{$\phi(a|x_t), B_a, x_t$}:
      \State $n(B_a)$ $\leftarrow$ $n(B_a)$ + $1$ 
      \State re$(B_a)$ $\leftarrow$ re$(B_a)$ + $\phi(a|x_t)$
      \If {conf($B_a$) $\le$ radius($B_a$)}     \hspace*{1.5em}//Activation
      \State $B'$ $\leftarrow$ $B(x_t, \frac{1}{2}{\rm radius}(B_a))$
      \State $\mathcal{A}_a \leftarrow \mathcal{A}_a \bigcup \{B'\}$
      \State $n(B')$ $\leftarrow$ $0$,  \  re$(B')$ $\leftarrow$ $0$
      \EndIf
      \EndFunction
  \end{algorithmic}
\end{algorithm}

\begin{remark}
We note that the index $I_t(B)$ defined above includes both the average reward $v_t(B)$ and a confidence radius, similar to the classic upper confidence bound (UCB) based approaches~\cite{auer2002finite}. Further, exploration is included in both probing and play stages. One may wonder if this is necessary since probing provides free observations, which may remove the necessity of exploration. However, as we show in our simulations, replacing $I_t(B)$ by $v_t(B)$ (so that no exploration is used) leads to suboptimal decisions. Intuitively, this is because although probing reduces uncertainty, it does not completely remove it for a small $K$. Thus, it is crucial to judiciously utilize the limited probing resource. 
\end{remark}

\subsection{Theoretical Analysis for CBwP with Bernoulli rewards}
In this section, we analyze the regret of Algorithm~\ref{alg:cbwp} in the special case when the rewards of arms follow Bernoulli distributions. 
\ignore{
Consider the optimal policy $\pi^\star$ defined in Lemma \ref{lemma:optimal-policy}.
For Bernoulli bandits, we have $G_{\pi_t}(x_t) =1-\prod\limits_{i=1}^{K+1}(1-\mu(a_{it}|x_t))$ for any $\pi_t$ that makes use of probing results, where $a_{it}$ is an arm probed or played in round $t$ chosen by $\pi_t$. Using mathematical induction, we can derive the following bound for the regret in each round. 
}

Before presenting our analysis, we first make the following claim. Proofs of these claims can be found in \cite{slivkins2011contextual}.

\begin{claim}
The following properties are maintained in every time round $t$:
\begin{enumerate}
\item For any active ball $B_a$,
{\rm conf}($B_a$) $\le$ $r(B_a)$ if and only if $B_a$ is a parent ball.
\item $\mathcal{A}_a$ covers the similarity space $(X,\mathcal{D})$. 
\item For any two active balls of same radius $r$ that are associated with the same arm, their centers are at distance at least $r$. 
\end{enumerate}
\label{claim1}
\end{claim}

\begin{claim}
If a ball $B_a$ is active in round $t$, then with probability at least $1-2T^{-3}$ we have that
\begin{align}
 {\rm |}v_t(B_a)-\mu(B_a)| \le r(B_a) + {\rm conf}_t(B_a)\label{eq:clean}
\end{align}
where $\mu(B_a)=\mu(a|x)$ and $x$ is the center of $B_a$. 
\label{claim2}
\end{claim}
We call a run of the algorithm {\it clean} if \eqref{eq:clean} is satisfied for every active ball and in every time round and {\it bad} otherwise. Since at most $K+1$ balls are activated in each round, the total number of active balls is bounded by $(K+1)T$. By applying the union bound, the probability that a bad run happens is at most $2(K+1)T^{-1}$. 

To derive the regret bound of our algorithm, we consider the optimal policy $\pi^\star$ defined in Lemma \ref{lemma:optimal-policy}. 
Consider any time round with context $x_t$. Let $a_{1t}^\star,a_{2t}^\star,\ldots,a_{(K+1)t}^\star$ denote the $K+1$ arms with largest $\mu(a|x_t)$ chosen by $\pi^\star$. 
Without loss of optimality, we assume that $a_{1t}^\star,a_{2t}^\star,\ldots,a_{(K+1)t}^\star$ are sorted non-increasingly by $I_t(B^{\rm{sel}}_{a_{it}^\star})$. Let $a_{it}$ denote the $i$-th ($i=1,...,K+1$) arm probed or played in time round $t$ in Algorithm~\ref{alg:cbwp}. For each probing step $i$ we have the following results. 

\begin{lemma}
Consider a clean run of the algorithm. We have
\begin{align}
 \Delta(a_{it}|x_t) &\triangleq \mu(a_{it}^\star|x_t) - \mu(a_{it}|x_t) \\
 &\le 14r(B^{\rm{sel}}_{a_{it}})
 \label{5}
\end{align}
\label{lem:4}
\end{lemma}

\begin{IEEEproof}
Fix a time round $t$ and consider any probing step $i$. We have
\begin{align*}
I_t(B^{\rm{sel}}_{a_{it}}) &\overset{(a)}{\ge} I_t(B^{\rm{sel}}_{a_{it}^\star})\\
&\overset{(b)}{=}r(B^{\rm{sel}}_{a_{it}^\star})+I_t^{\rm pre}(B_{a_{it}^\star}')+\mathcal{D}(B^{\rm{sel}}_{a_{it}^\star}, B_{a_{it}^\star}')\\
&\overset{(c)}{\ge} r(B^{\rm{sel}}_{a_{it}^\star})+\mu(B'_{a_{it}^\star}) +\mathcal{D}(B^{\rm{sel}}_{a_{it}^\star}, B_{a_{it}^\star}')\\
&\overset{(d)}{\ge} r(B^{\rm{sel}}_{a_{it}^\star})+\mu(B^{\rm{sel}}_{a_{it}^\star})\\
&\overset{(e)}{\ge} \mu(a_{it}^\star|x_t)
\end{align*}
where (a) is due to the probing and playing rules, (b) follows from Eq.~(\ref{UCB}) for some active ball $B_{a_{it}^\star}'$, (c) follows from Eq. (\ref{preUCB}) and the clean run assumption (\ref{eq:clean}), (d) holds from the Lipschitz property (\ref{Lipschitz}), and (e) holds from the Lipschitz property (\ref{Lipschitz}) and the fact that $x_t \in B^{\rm{sel}}_{a_{it}^\star}$.
Let $B_{a_{it}}^{\rm par}$ be the parent of $B_{a_{it}}^{\rm{sel}}$, and by the activation rule, we have
\begin{align}
{\rm max}(\mathcal{D}(B_{a_{it}}^{\rm sel}, \ B_{a_{it}}^{\rm par}), {\rm conf}_t(B_{a_{it}}^{\rm par}))\le r(B_{a_{it}}^{\rm par})
\label{parent}
\end{align}
It follows that 
\begin{align*}
I_t^{\rm pre}(B_{a_{it}}^{\rm par}) &\overset{(a)}{=} v_t(B_{a_{it}}^{\rm par}) + r(B_{a_{it}}^{\rm par}) + {\rm conf}_t(B_{a_{it}}^{\rm par}) \\
&\overset{(b)}{\le}\mu(B_{a_{it}}^{\rm par}) + 2r(B_{a_{it}}^{\rm par})+2{\rm conf}_t(B_{a_{it}}^{\rm par})\\
&\overset{(c)}{\le}\mu(B_{a_{it}}^{\rm par}) + 4r(B_{a_{it}}^{\rm par})\\
&\overset{(d)}{\le}\mu(B_{a_{it}}^{\rm sel}) + 5r(B_{a_{it}}^{\rm par}) \numberthis \label{eq:bound-par}
\end{align*}
where (a) is due to Eq.~(\ref{preUCB}), (b) follows from the clean run assumption (\ref{eq:clean}), (c) follows from Eq.~(\ref{parent}), and (d) holds from (\ref{parent}) and the Lipschitz property (\ref{Lipschitz}).
Now we can upper bound $I_t(B_{a_{it}}^{\rm sel})$:
\begin{align*}
I_t(B_{a_{it}}^{\rm sel}) &\overset{(a)}{\le} r(B_{a_{it}}^{\rm sel})+I_t^{\rm pre}(B_{a_{it}}^{\rm par})+\mathcal{D}(B_{a_{it}}^{\rm sel}, B_{a_{it}}^{\rm par})\\
&\overset{(b)}{\le}r(B_{a_{it}}^{\rm sel})+I_t^{\rm pre}(B_{a_{it}}^{\rm par})+r(B_{a_{it}}^{\rm par})\\
&\overset{(c)}{\le}r(B_{a_{it}}^{\rm sel})+\mu(B_{a_{it}}^{\rm sel}) + 6r(B_{a_{it}}^{\rm par})\\
&\overset{(d)}{=}\mu(B_{a_{it}}^{\rm sel})+13r(B_{a_{it}}^{\rm sel})\\
&\overset{(e)}{\le}\mu(a_{it}|x_t)+14r(B_{a_{it}}^{\rm sel})  \numberthis \label{eq:bound-sel}
\end{align*}
where (a) is due to Eq.~(\ref{UCB}), (b) follows from Eq.~(\ref{parent}), (c) follows from Eq.~\eqref{eq:bound-par}, 
(d) is due to the fact that $r(B_{a_{it}}^{\rm par})=2r(B_{a_{it}}^{\rm sel})$, and (e) holds from the Lipschitz property (\ref{Lipschitz}).
From the above results, we can get
\begin{align}
\mu(a_{it}^\star|x_t) \le I_t(B_{a_{it}}^{\rm sel}) \le \mu(a_{it}|x_t) + 14r(B_{a_{it}}^{\rm sel})
\end{align}
So finally have
\begin{align*}
\Delta(a_{it}|x_t) &\triangleq \mu(a_{it}^\star|x_t) - \mu(a_{it}|x_t) \\
&\le 14r(B_{a_{it}}^{\rm sel}).
\end{align*}
\end{IEEEproof}

\begin{corollary}
Consider a clean run of our algorithm. If ball $B_a$ is activated in round $t$ and $B_{a_{it}}^{\rm sel}$ is the parent of $B_a$. then we have
$\Delta(a_{it}|x_t) \le 12r(B_a)$.
\label{cor}
\end{corollary}
\begin{IEEEproof}
As $B_{a_{it}}^{\rm sel}$ is the parent of $B_a$, by Lemma~\ref{lem:4} and the activation rule we have $\Delta(a_{it}|x_t)\le 14r(B_{a_{it}}^{\rm sel})=28r(B_a)$.
To prove the corollary, we replace \eqref{eq:bound-sel} in the proof of Lemma~\ref{lem:4} to show that 
if $B_{a_{it}}^{\rm sel}$ is a parent ball, then $\Delta(a_{it}|x_t)\le 6r(B_{a_{it}}^{\rm sel})$. This can be shown as follows:
\begin{align*}
I_t(B_{a_{it}}^{\rm sel}) &\overset{(a)}{\le} r(B_{a_{it}}^{\rm sel})+I_t^{\rm pre}(B_{a_{it}}^{\rm sel})\\
&\overset{(b)}{=}v_t(B_{a_{it}}^{\rm sel}) + 2r(B_{a_{it}}^{\rm sel}) + {\rm conf}_t(B_{a_{it}}^{\rm sel})\\
&\overset{(c)}{\le}\mu(B_{a_{it}}^{\rm sel})+3r(B_{a_{it}}^{\rm sel})+ 2{\rm conf}_t(B_{a_{it}}^{\rm sel})\\
&\overset{(d)}{\le}\mu(a_{it}|x_t)+6r(B_{a_{it}}^{\rm sel})
\end{align*}
where (a) is due to Eq.~\eqref{UCB}, (b) follows from Eq.~\eqref{preUCB}, (c) follows from the clean run assumption (\ref{eq:clean}), and (d) holds from the Lipschitz property (\ref{Lipschitz}) and the activation rule. The rest of the proof follows the same reasoning as in Lemma~\ref{lem:4}.

\end{IEEEproof}

\begin{lemma}
For any round $t$, we have 
\begin{align*}
\Delta(G_{\pi_t}|x_t))\triangleq & G_{\pi^\star}(x_t)-G_{\pi_t}(x_t) \\
\le & (1-\mathop{\rm min}\limits_{i\in{\{1,2,\ldots,K\}}}\mu(a_{it}|x_t))^K\sum\limits_{j=1}^{K+1}\Delta(a_{jt}|x_t).
\end{align*}
\label{lemma5}
\end{lemma}


\begin{IEEEproof}
Consider any round $t$. Recall that $a_{1t}^\star, a_{2t}^\star,\ldots,$ $a_{(K+1)t}^\star$ are sorted non-increasingly by $I_t(B^{\rm{sel}}_{a_{it}^\star})$. We have
\begin{align*}
G_{\pi^\star}(x_t) &=1-\prod\limits_{i=1}^{K+1}(1-\mu(a_{it}^\star|x_t)), \\
G_{\pi_t}(x_t) &=1-\prod\limits_{i=1}^{K+1}(1-\mu(a_{it}|x_t)).
\end{align*}
To simplify the notation, let $u_i \triangleq 1-\mu(a_{it}|x_t)$, $v_i \triangleq 1-\mu(a^*_{it}|x_t)$, and $\delta_i \triangleq \Delta(a_{it}|x_t)$. We have
\begin{align*}
\Delta(G_{\pi_t}|x_t)) = &G_{\pi^\star}(x_t)-G_{\pi_t}(x_t)\\
                     =&\prod\limits_{i=1}^{K+1}u_i-\prod\limits_{i=1}^{K+1}v_i \\
=&\prod\limits_{i=1}^{K+1}(v_i+\delta_i)-\prod\limits_{i=1}^{K+1}v_i
\end{align*}
We claim that 
\begin{align*}
\prod\limits_{i=1}^{K+1}(v_i+\delta_i)=&\prod\limits_{i=1}^{K+1}v_i+\sum\limits_{j=1}^{K+1} \left( \prod\limits_{i=1}^{j-1}v_i \prod\limits_{i=j+1}^{K+1}u_i\right) \delta_j.
\label{math}
\end{align*}
We prove the claim by induction. Let $L_K$ denote the left side of the equation. First when $K=1$, we have
\begin{align*}
L_1=\prod\limits_{i=1}^{2}(v_i+\delta_i)=&v_{1}v_{2}+u_{2}\delta_1+v_{1}\delta_2\\
=&\prod\limits_{i=1}^{2}v_i+\prod\limits_{i=2}^{2}u_i\delta_1+\prod\limits_{i=1}^{1}v_i\delta_{2}.
\end{align*}
So the claim holds when $K=1$.
Assume it holds when $K=m$, we have
\begin{align*}
L_{m+1} =& L_{m} (v_{m+2}+\delta_{m+2}) \\
\overset{(a)}=&\left(\prod\limits_{i=1}^{m+1}v_i+\sum\limits_{j=1}^{m+1} \left( \prod\limits_{i=1}^{j-1}v_i \prod\limits_{i=j+1}^{m+1}u_i\right) \delta_j \right)(v_{m+2}+\delta_{m+2}) \\
\overset{(b)}=&\left(\prod\limits_{i=1}^{m+1}v_i\right) v_{m+2} + \left(\prod\limits_{i=1}^{m+1}v_i\right) \delta_{m+2} \\ &\hspace{5ex}+\left(\sum\limits_{j=1}^{m+1} \left( \prod\limits_{i=1}^{j-1}v_i \prod\limits_{i=j+1}^{m+1}u_i\right) \delta_j \right) u_{m+2} \\
=&\prod\limits_{i=1}^{m+2}v_i + \left(\prod\limits_{i=1}^{m+1}v_i\right) \delta_{m+2} \\ &\hspace{5ex}+\sum\limits_{j=1}^{m+1} \left( \prod\limits_{i=1}^{j-1}v_i \prod\limits_{i=j+1}^{m+2}u_i\right) \delta_j \\
=&\prod\limits_{i=1}^{m+2}v_i + \sum\limits_{j=1}^{m+2} \left( \prod\limits_{i=1}^{j-1}v_i \prod\limits_{i=j+1}^{m+2}u_i\right) \delta_j \\
\end{align*}
where (a) follows from the inductive hypothesis and (b) holds because $u_i=v_i+\delta_i$. Therefore, the claim holds for any $K \geq 1$.

Now we have
\begin{align*}
\Delta(G_{\pi}|x_t))&=\prod\limits_{i=1}^{K+1}(v_i+\delta_i)-\prod\limits_{i=1}^{K+1}v_i\\
&=\sum\limits_{j=1}^{K+1} \left( \prod\limits_{i=1}^{j-1}v_i \prod\limits_{i=j+1}^{K+1}u_i\right) \delta_j \\
&\leq\sum\limits_{j=1}^{K+1} \left( \prod\limits_{i=1, i \neq j}^{K+1}u_i\right) \delta_j \\
&\le(1-\mathop{\rm min}\limits_{i\in{\{1,2,\ldots,K\}}}\mu(a_{it}|x_t))^K\sum\limits_{j=1}^{K+1}\delta_j
\end{align*}

\end{IEEEproof}

\ignore{To bound the total regret, the main idea is to show that with high probability, (1) $\Delta(a_{it}|x_t)$ is bounded by  $r(B^{\rm{sel}}_{a_{it}})$ times a constant for any $a_{it}$ and $x_t$, and (2) the total number of balls associated with any given arm that have radius $r$ and have been activated throughout the execution of the algorithm is bounded by $N_{r}$, the $r$-packing number\cite{slivkins2011contextual}. We then have the following main result. The detailed proof can be found in our online technical report~\cite{tech}.}

We then bound the total regret. For a radius $r=2^{-j}$ for some $j\in\mathbb{N}$, we define $F_{ar}$ as the set of balls associated with arm $a$ with radius $r$ that have been activated throughout the execution of the algorithm. In addition, for each ball $B_a\in F_{ar}$, we define $S_{B_a}$ as the set of rounds including (1) the round when $B_a$ was activated and (2) all rounds $t$ when $B_a$ was probed or played and was not a parent ball. From the activation rule and Eq.~\eqref{eq:conf}, we have
\begin{equation}
\label{eq:sba1}
|S_{B_a}| \le 16r^{-2}(B_a){\rm log}T
\end{equation}
By Claim \ref{claim1}.3, the centers of balls in $F_{ar}$ are within distance at least $r$ from one another. So $|F_{ar}|\le N_{r}$, where $N_{r}$ is the $r$-packing number\cite{slivkins2011contextual}. Fix some $r_0\in(0,1)$. In each round $t$ when a ball of radius $< r_0$ associated with arm $a_{it}$  was selected or activated, the regret $\Delta(a_{it}|x_t)\le 14r_0$ according to Lemma~\ref{lem:4} and Corollary~\ref{cor}. 

We then have the following main result. 

\begin{figure*}[t]
 	\vspace{-10pt}
 	\centering
 		\subfloat[]{%
 		\includegraphics[width=0.21\textwidth]{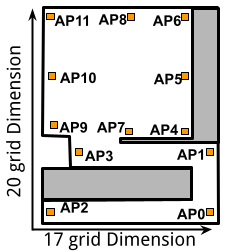}%
 		\label{student center}
 	}
 	\subfloat[]{%
 		\includegraphics[width=0.25\textwidth]{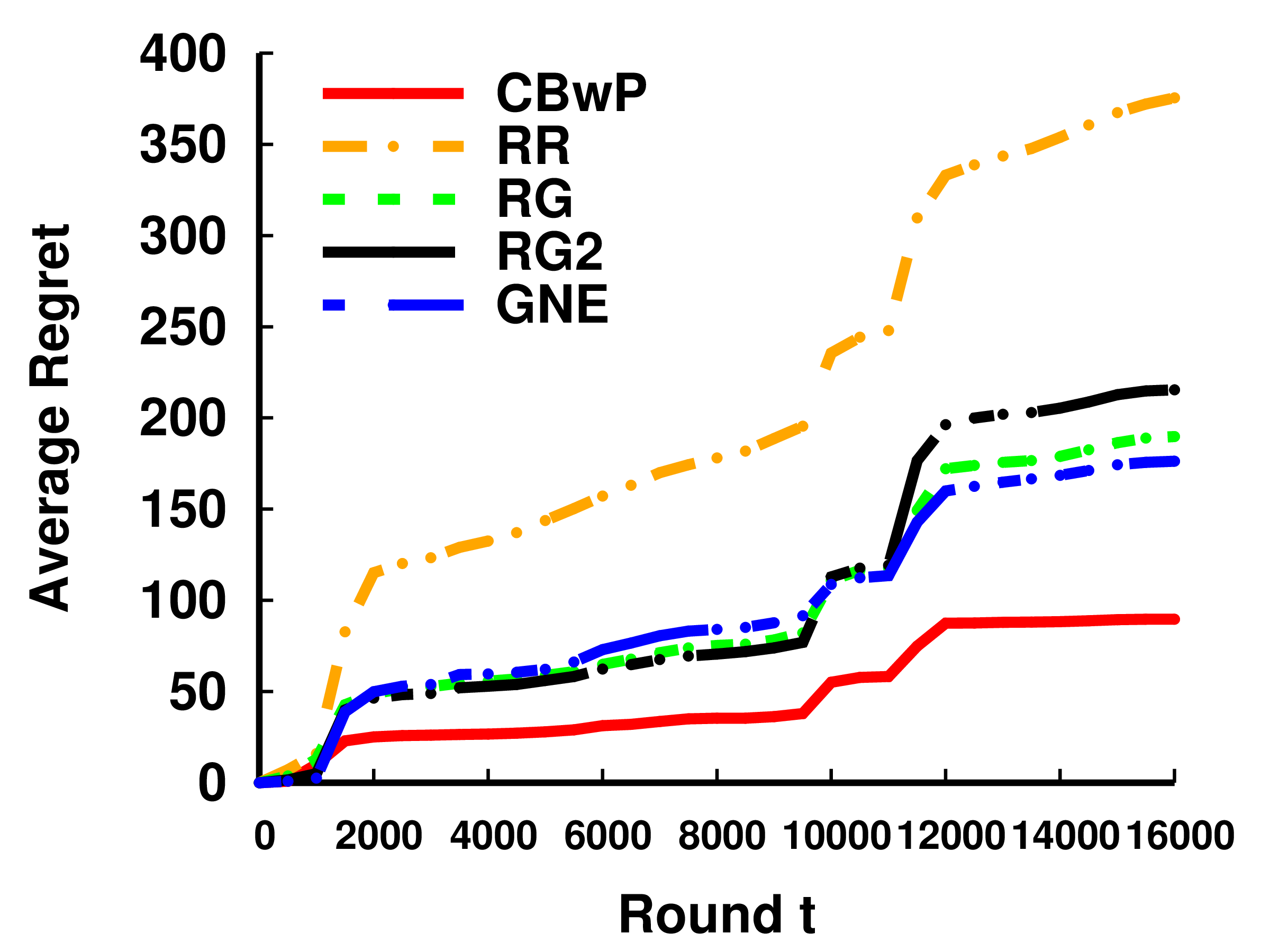}%
 		\label{fig1}
 	}
 	\subfloat[]{%
 		\includegraphics[width=0.25\textwidth]{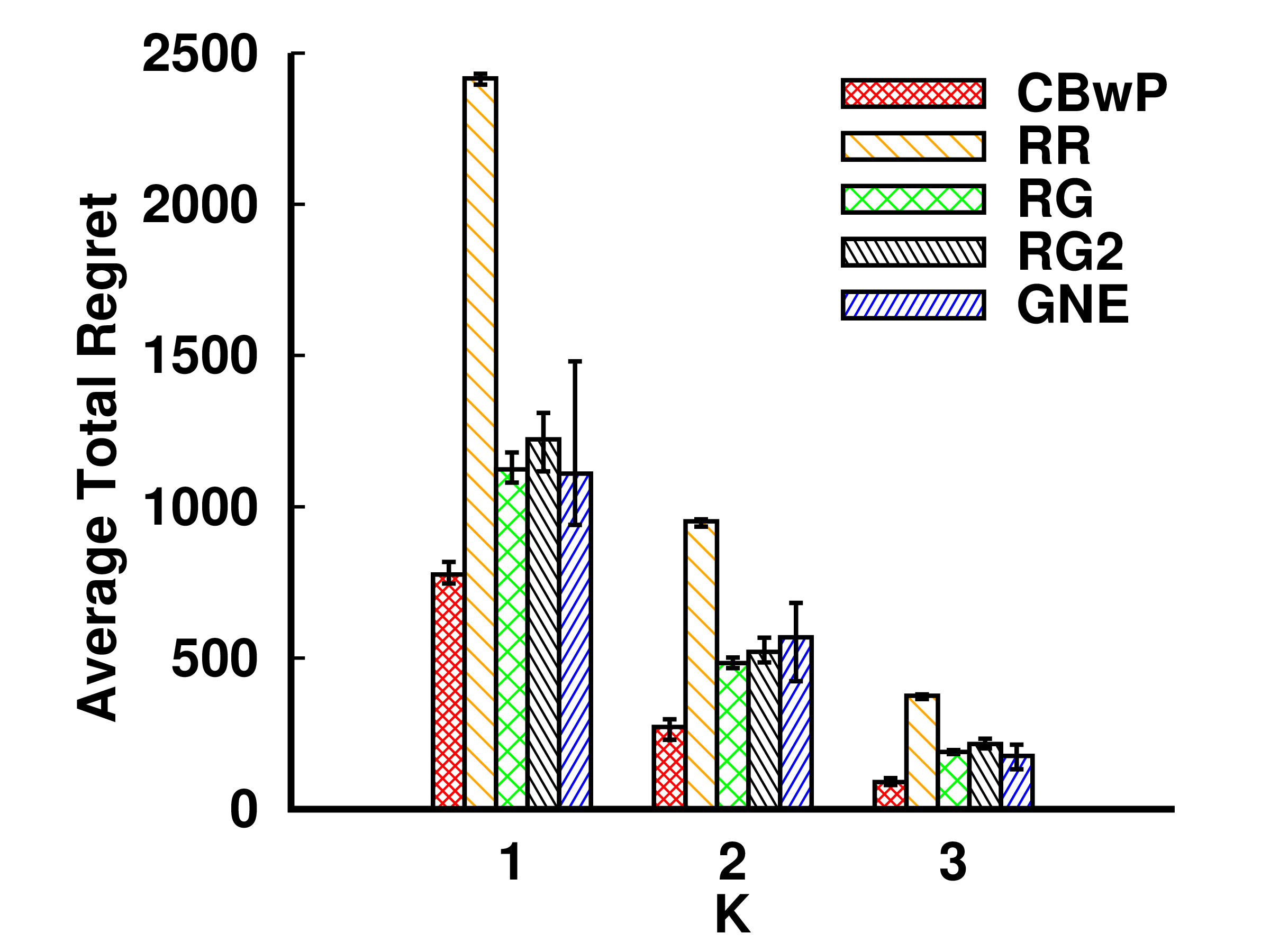}%
 		\label{fig2}
 	}
 	\subfloat[]{%
 		\includegraphics[width=0.25\textwidth]{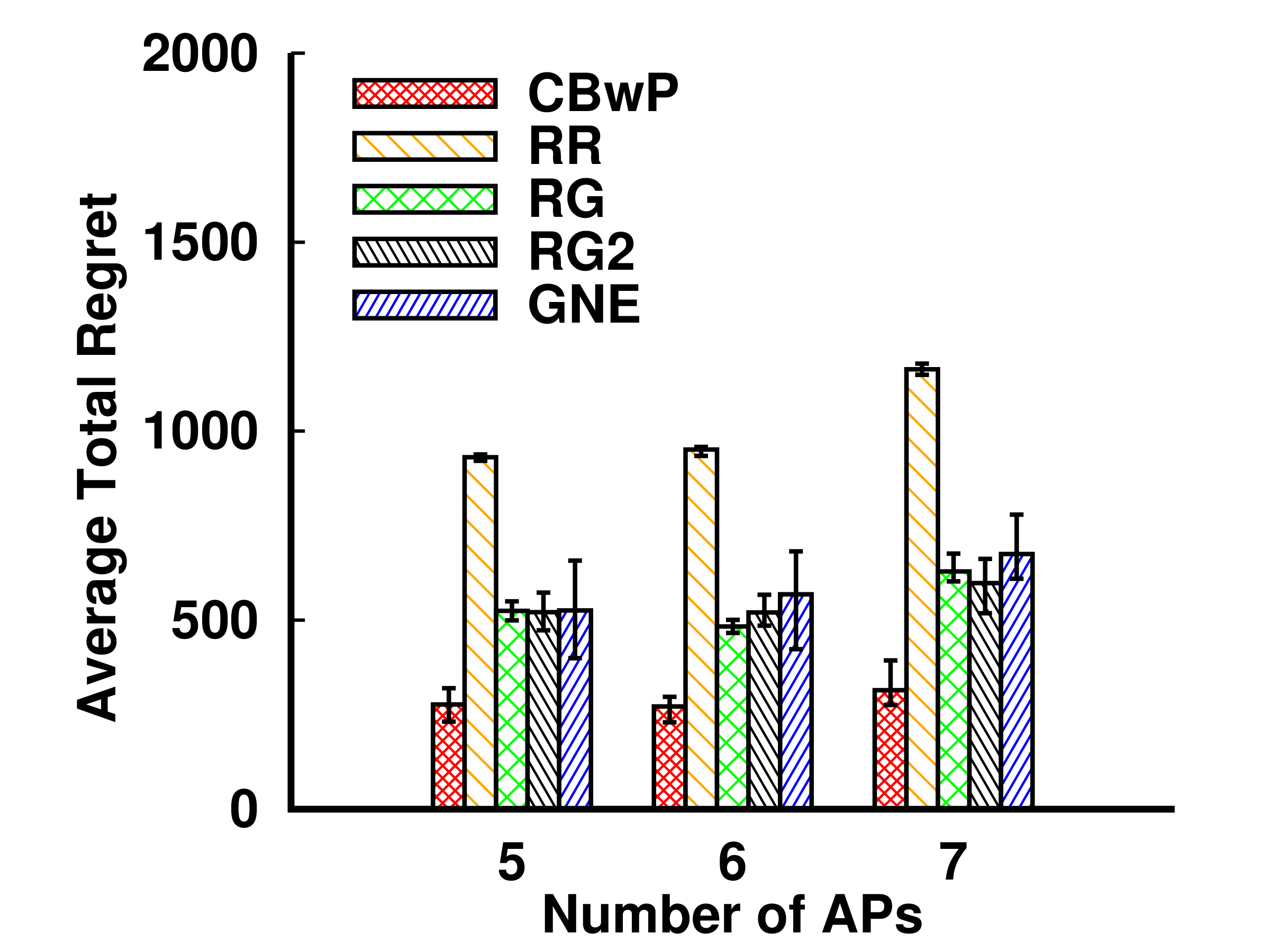}%
 		\label{fig3}
 	} 	\vspace{-5pt}

 	 \caption{\small (a) Scenario: Student Hall; (b)-(d) Comparing CBwP with baselines on average regret: (b) $K = 3, N = 6$; (c) $N=6$; (d) $K=2$.}
 	\vspace{-15pt}
 	\label{fig}
 \end{figure*}
\begin{theorem} The total expected regret of Algorithm~\ref{alg:cbwp} is bounded as follows:
\begin{align*}
\mathbb{E}(R(T)){\le} & (K+1)(14r_0HT+2) \\
&+224NH \left(\sum_{r=2^{-j}: r_0\le r \le 1}r^{-1}N_{r}\right){\log}T 
\end{align*}
where $H \triangleq (1-\mathop{\rm min}\limits_{a \in A, x \in X}\mu(a|x))^K$. 

\label{theorem}
\end{theorem}

\begin{IEEEproof}
For any clear run, we have
\begin{align*}
R(T)=&\sum_{t=1}^T (G^{\star}(x_t) - G_{\pi_t}(x_t))\\
\overset{(a)}{\le} &\sum_{t=1}^{T}H\sum\limits_{j=1}^{K+1}\Delta(a_{jt}|x_t) \\
\overset{(b)}\le& 14r_0(K+1)HT\\
&\hspace{5ex}+\sum_{r=2^{-j}: r_0\le r \le 1}\sum_{a \in A}\sum_{B_a\in F_{ar}}\sum_{t\in S_{B_a}}H\Delta(a_{it}|x_t) \\ 
\overset{(c)}{\le} & 14r_0(K+1)HT \\
&\hspace{5ex}+\sum_{r=2^{-j}: r_0\le r \le 1}\sum_{a \in A}\sum_{B_a\in F_{ar}}|S_{B_a}|H14r(B_a)\\
\overset{(d)}{\le} & 14r_0(K+1)HT \\
&\hspace{0ex}+H\sum_{r=2^{-j}: r_0\le r \le 1}\sum_{a \in A}\sum_{B_a\in F_{ar}}16r^{-2}(B_a){\rm log}T14r(B_a)\\
{\le} & 14r_0(K+1)HT +224H\sum_{r=2^{-j}: r_0\le r \le 1}\sum_{a \in A}r^{-1}N_{r}{\rm log}T\\
{\le} &  14r_0(K+1)HT+224NH\sum_{r=2^{-j}: r_0\le r \le 1}r^{-1}N_{r}{\rm log}T \numberthis \label{eq:clean-bound}
\end{align*}
where (a) follows from Lemma \ref{lemma5}, (b) and (c) are due to Lemma~\ref{lem:4} and Corollary~\ref{cor}, and (d) holds from Eq.~\eqref{eq:sba1}.

For any bad run, we trivially have $R(T)\leq T$. Let $Z(T)$ denote the right side of~\eqref{eq:clean-bound}. The total expected regret can be bounded as:
\begin{align*}
\mathbb{E}(R(T)) &= \mathbb{E}(R(T)|\textrm{clean run})\textrm{Pr(clean run)} \\
&\hspace{10ex}+\mathbb{E}(R(T)|\textrm{bad run})\textrm{Pr(bad run)}\\
&\leq Z(T)+T\times2(K+1)T^{-1}\\
&\leq Z(T)+2(K+1)
\end{align*}
which gives us the desired bound. 
\end{IEEEproof}


Note that the bound in the theorem can be tightened by taking inf on all $r_0\in(0,1)$.
\section{Evaluation}

In this section, we present the evaluation results of CBwP by comparing it with four baselines: (a) RR (random probing and random play): randomly probing $K$ arms and then randomly playing an arm (if none of the probed arms gives a reward of 1); (b) RG (random probing and greedy play with exploration): randomly probing $K$ arms and then playing the arm $a$ with maximum $v(a)$ (as defined in the playing rule in Section~\ref{sec:algo}); (c) RG2 (random probing and greedy play without exploration): randomly probing $K$ arms and then playing the arm $a$ with maximum $v'(a)$, where $v'(a) = \phi(a|x_t)$ if $a$ has been probed and $v'(a) = v_t(B^{\rm{sel}}_{a})$ otherwise; 
(d) GNE (greedy probing and play without exploration): this is a variant of Algorithm~\ref{alg:cbwp} where we replace the index of a ball $I_t(B)$ with its average reward $v_t(B)$ in both the probing the playing rules.

\subsection{Evaluation Settings}

In order to evaluate the system, we collect the channel traces from the real testbeds with 802.11ad routers and laptops, and a commercial mmWave channel simulator (Remcom Insite \cite{remcom}). We collected SNR traces in a student hall (Scenario 1) with real testbeds at 250 different locations at the granularity of 0.8m. 12 Airfide\cite{airfide} 802.11ad APs are deployed in the student hall and each of them is equipped with 8 phased array antennas with 64 sectors. We use the Acer TravelMate-648 laptops with a single phased array and 36 sectors as the clients. We modified the open-source driver of both devices to extract the SNR and beamforming information.

After getting the signal strength channel traces from our testbed and the Remcom channel simulator, we use MCS-SNR table from the 802.11ad \cite{80211ad} to map the SNR to get the average throughput of the link (as in \cite{mobihoc18-mmchoir}) based on the best Tx/Rx sector pair of the link found using the beamforming process. We normalize the average throughput of each AP $a$ as $\mu(a|x)$ for each context $x$ (location of a client). 
We consider Bernoulli distribution with probability $\mu(a|x)$ to get reward $1$ for each time round.

For the mobility traces, we select from 15 to 80 clients randomly located 
where each client follows a specific walking pattern. We set the walking pattern of the clients by observing the typical walking behaviors in each room. 
We assume the walking speed as $1 m/s$ for all clients and for each time slot, the client will locate at one of the grids where the channel is measured by our testbed or the channel simulator as described before. In the simulations, we choose 10 clients' traces where each of the traces has 80 steps and each step includes 20 time rounds.

For computing the distance $\mathcal{D}(x,x')$ between two arbitrary locations $x$ and $x'$, 
we let $\mathcal{D}(x,x')=E(x,x')/{\rm Dia}$ where $E(x,x')$ is the Euclidean distance between $x$ and $x'$ and ${\rm Dia}$ is the longest diagonal line length. 

We note that although we consider an indoor environment in the evaluation, our approach can be readily applied to outdoor settings as well, e.g., mmWave based vehicular networks~\cite{fml2018}. 
\subsection{Results}

We evaluate all the algorithms 
with different $K$ and number of APs $N$. 
For each $N$, we randomly pick 5 different sets of APs with size $N$. In each setting, we run all the algorithms 10 times (with the same random seeds) and take the average. 

Fig.~\ref{fig1} shows how the average total expected regret changes with time. 
Compared with the baselines, CBwP's regret increases more slowly and after time round 1,500, the average regret of CBwP's is always lower than the others. In the figure, there are two jumps around time rounds 9,500 and 11,000, respectively, which correspond to the starting points of  two new clients who entered the room from the locations less explored in previous time rounds, e.g., the bottom area where most APs are blocked. 

Fig.~\ref{fig2} shows the average total expected regret of all the five algorithms under different $K$ with $N=6$. We observe that 
CBwP outperforms all the baselines irrespective of $K$. 
In addition, with the value of $K$ increasing, the average expected regret gradually declines in all the algorithms. This is expected as a larger $K$ provides a higher chance of finding a good arm to play. 
Fig.~\ref{fig3} shows the average total expected regret of the five algorithms for different numbers of APs with $K=2$. 
CBwP again outperforms all the baselines. Further, with AP increasing, the average expected regret gradually rises for the baselines, indicating their difficulty of scaling to more APs. In contrast, the performance of CBwP is stable.

\section{Conclusion}
In this paper, we consider the problem that APs cooperatively serve a mobile client with unknown date rates. We propose contextual bandit with probing (CBwP) as a novel bandit learning framework that incorporates joint probing and play to solve this problem. We derive structural properties of the optimal offline solution and an efficient online learning algorithm to CBwP. We further establish the regret bound of CBwP for links with Bernoulli data rates. Our CBwP model is a novel extension to the classic contextual bandit model and can potentially be applied to a large class of sequential decision-making problems that involve joint probing and play under uncertainty.

\section*{Acknowledgment}
This work was supported in part by NSF grant CNS-1816943.

\vspace{12pt}

\bibliographystyle{ieeetr} 
\bibliography{references}
\end{document}